\documentclass[10pt,twocolumn,letterpaper]{article}

\usepackage{cvpr}              

\usepackage[dvipsnames]{xcolor}

\definecolor{cvprblue}{rgb}{0.21,0.49,0.74}
\usepackage[pagebackref,breaklinks,colorlinks,citecolor=cvprblue]{hyperref}
\usepackage[accsupp]{axessibility} 

\usepackage{multirow}
\usepackage{color, colortbl}
\definecolor{gray}{gray}{0.85}

\def\paperID{XXX} 
\def\confName{CVPR}
\def\confYear{2024}

\begin{document}
\title{Backpropagation-free Network for 3D Test-time Adaptation}

\author{Yanshuo Wang$^{1,2}$, 
Ali Cheraghian$^{2}$, 
Zeeshan Hayder$^{2}$,
Jie Hong$^{1,2}$\thanks{Corresponding author},
Sameera Ramasinghe$^{5}$,
\and
Shafin Rahman$^{3}$,
David Ahmedt-Aristizabal$^{2}$,
Xuesong Li$^{1,2}$,
Lars Petersson$^{2}$, 
Mehrtash Harandi$^{2,4}$ \\
$^{1}$Australian National University, 
$^{2}$Data61-CSIRO, Australia \\ 
$^{3}$North South University, Bangladesh, 
$^{4}$Monash University, Australia \\
$^{5}$Amazon, Australia \\
{\tt\small \{yanshuo.wang, jie.hong, xuesong.li\}@anu.edu.au}, \\
{\tt\small \{ali.cheraghian, zeeshan.hayder, david.ahmedtaristizabal, lars.petersson\}@data61.csiro.au}, \\
{\tt\small shafin.rahman@northsouth.edu, mehrtash.harandi@monash.edu, ramasisa@amazon.com}
}
\maketitle

\begin{abstract}
Real-world systems often encounter new data over time, which leads to experiencing target domain shifts. Existing Test-Time Adaptation (TTA) methods tend to apply computationally heavy and memory-intensive backpropagation-based approaches to handle this. Here, we propose a novel method that uses a backpropagation-free approach for TTA for the specific case of 3D data. Our model uses a two-stream architecture to maintain knowledge about the source domain as well as complementary target-domain-specific information. The backpropagation-free property of our model helps address the well-known forgetting problem and mitigates the error accumulation issue. The proposed method also eliminates the need for the usually noisy process of pseudo-labeling and reliance on costly self-supervised training. Moreover, our method leverages subspace learning, effectively reducing the distribution variance between the two domains. Furthermore, the source-domain-specific and the target-domain-specific streams are aligned using a novel entropy-based adaptive fusion strategy. Extensive experiments on popular benchmarks demonstrate the effectiveness of our method. The code will be available at \url{https://github.com/abie-e/BFTT3D}.
\end{abstract}
\section{Introduction}
\label{sec:intro}

\begin{figure}[t]\centering
\includegraphics[width=1.0\linewidth,trim=0cm 0cm 0cm 0cm, clip]{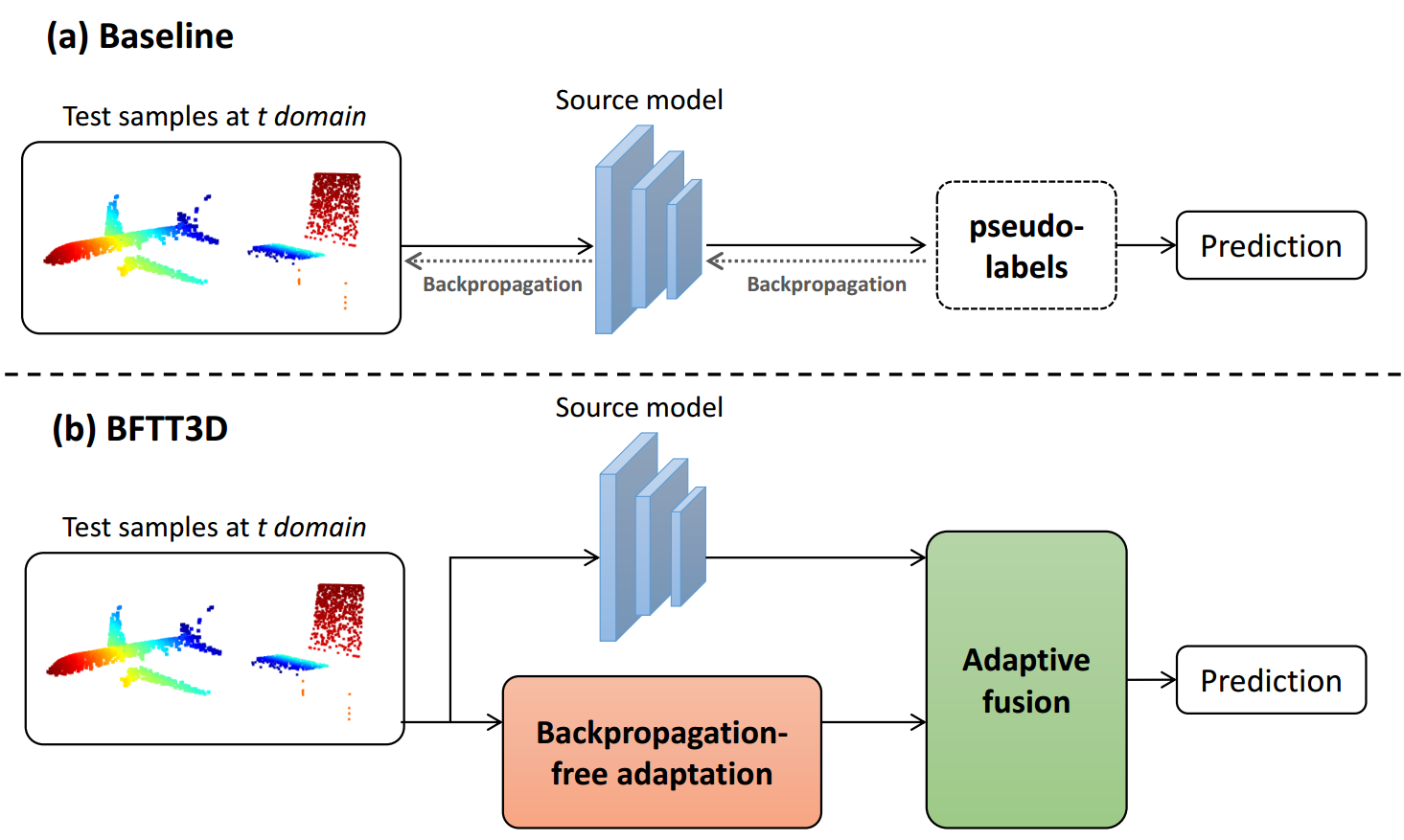}
\caption{(a) Baseline. When faced with new point cloud samples at test time $t$, most existing methods generate the pseudo-labels and train the source model in a self-supervised manner. (b) Backpropagation-free test-time 3D model (BFTT3D). BFTT3D adopts a backpropagation-free adaptation module to output the target-specific logit, which fuses with the logit from the source model for prediction. Compared to the baseline, BFTT3D does not require any pseudo-labeling process and backpropagation.}
\label{fig:proj_issue}
\end{figure}

In recent years, 3D point cloud processing has experienced significant growth~\cite{qi2017pointnet, pointnet++, dgcnn, Xiang_2021_ICCV}, driven largely by advances in deep learning techniques. While researchers have made commendable contributions in these areas, their focus has predominantly been on controlled environments. A notable real-world scenario is Test-Time Adaptation (TTA),  
which has recently gained substantial attention from researchers~\cite{gandelsman2022test,sun2020test,liu2021ttt++,iwasawa2021test,gao2022back,lim2023ttn,yeo2023rapid,niu2023towards,schneider2020improving}, owing to its critical applications in practical settings. In TTA, the model must rapidly adapt to new test samples during testing to provide accurate final predictions. For example, consider a self-driving car equipped with a lidar object detector designed to locate humans and cars on the street. Although this detector may have been trained exclusively under ideal weather conditions, it must swiftly adapt to new environments during test time, such as snowy or rainy weather conditions. This highlights the pressing need to adopt TTA to ensure the car can adapt swiftly to these dynamic environments without extensive retraining. In this paper, we endeavor to propose an innovative and efficient method for TTA on 3D data~\cite{hatem2023point, mirza2023mate}, an area that has received comparatively less attention in the literature compared to 2D images~\cite{chen2022contrastive,wang2021tent,liang2020we, li2016revisiting}. Nonetheless, from a problem-solving standpoint, it holds equal significance for the computer vision community.

In a TTA scenario, the model must swiftly adapt to a new target domain on the fly, using unlabeled test samples. It accomplishes this task without direct access to the original training data, instead relying on a pre-trained model derived from it. Traditionally, two prominent techniques are employed to tackle the TTA challenge. In the first category, the model utilizes a teacher-student architecture~\cite{wang2022continual, dobler2022robust, pan2023ssfe} to generate pseudo-labels for given test samples, enabling it to adjust to a new domain. The model is then updated through back-propagation using these pseudo-labels. While this strategy leverages pseudo-labels effectively, relying solely on them can introduce noisy supervision, leading to the accumulation of errors from previous test samples and significantly hindering performance on subsequent new test samples. Moreover, it may result in losing valuable knowledge from the source domain during adaptation to a new one due to the forgetting issue. In the second category, self-supervised methods~\cite{mirza2023mate} are employed to adapt the model to new domains. Although they perform well, they tend to be slower, making them less suitable for real-time applications where rapid adaptation is crucial. In this paper, we introduce a novel approach for 3D test-time learning that overcomes forgetting and error accumulation issues and allows for fast adaptation. This makes it a promising choice for addressing real-world application problems, which is pivotal in test-time learning.

In this paper, as shown in Figure~\ref{fig:proj_issue}, we introduce a novel approach, backpropagation-free test-time 3D model (BFTT3D), for TTA that circumvents the reliance on pseudo-labels and avoids needing a complex and sluggish self-supervised learning technique. 
It operates without necessitating updates to the model parameters during adaptation, thus sidestepping the issue of error accumulation. To delve deeper into our proposed method, we leverage an existing 3D point cloud model~\cite{qi2017pointnet,pointnet++,dgcnn,Xiang_2021_ICCV} as the source model, meticulously trained on the source data. The source model crafts a source-domain-specific (or source-specific) feature description when presented with a test sample from a target domain. The source model is kept frozen during upcoming test samples, which helps us to keep useful knowledge of the source domain. Concurrently, we generate a target-domain-specific (or target-specific) description tailored to the given test samples from the target domain. Within our designed backpropagation-free adaptation module, our focus lies in narrowing the domain gap between the emerging target domain and the established source domain within a defined subspace, effectively minimizing the distribution divergence between the two. Notably, our dynamic domain adaptation methods enable seamless adjustment to forthcoming test samples from the target domain, rendering them highly versatile in TTA scenarios. In the final stage, we integrate the information from both source-specific and target-specific information, leveraging the entropy data from both models during adaptation. Our proposed solution is highly adaptive as it can grow with the data, accommodating new information without requiring updates to predefined parameters (avoiding the computational cost of backpropagation), and well-suited for scenarios where data is constantly changing or expanding.

In summary, the main contributions of our paper are as follows:
\begin{itemize}
\item We present a novel and efficient approach, BFTT3D, for 3D TTA, eliminating the necessity for extensive backpropagation. This approach is less susceptible to the impact of noisy supervision derived from pseudo-labels and does not entail parameter fine-tuning during adaptation, consequently sidestepping error accumulation and forgetting issues. 

\item The backpropagation-free adaptation is proposed along with the source model to enhance the model's adaptability to specific domains. It is introduced without adding the extra model parameters that need backpropagation.

\item Using entropy information, we fuse the source-specific feature from the source model and the target-specific feature from the backpropagation-free adaptation. Like the backpropagation-free adaptation module, the fusion module does not introduce extra model parameters that need backpropagation.

\item The experimental outcomes demonstrate the superiority of the proposed BFTT3D on popular benchmarks, including including ModelNet-40C~\cite{sun2022benchmarking} and ScanObjectNN-C~\cite{mirza2023mate}. The algorithm is validated to be effective and efficient in 3D TTA.
\end{itemize}
\section{Related work}
\label{sec:related_work}

\noindent\textbf{Point cloud domain adaptation:}
In recent years, 3D sensors have become integral components of perception systems. The field of 3D point cloud processing has experienced remarkable growth, primarily fueled by advancements in deep learning techniques~\cite{qi2017pointnet, pointnet++, dgcnn, Xiang_2021_ICCV, hong2023pointcam, li2019three}.
Unsupervised domain adaptation (UDA) is increasingly prominent in the 3D vision for mitigating domain gaps between source and target datasets without target label information. 
UDA methods can be categorized into two groups. The first group requires labeled source data and source-trained models for adaptation to the target domain, while the second group employs source-free domain adaptation approaches.
Wang \textit{et al.}~\cite{wang2020train} introduce a statistic normalization approach to handle the gaps. Many of the following works leverage self-training to resolve the domain adaptation problem. These methods make improvements mainly on obtaining pseudo labels of higher quality \textit{e.g.}~with a memory bank (ST3D)~\cite{yang2021st3d} or by harnessing scene flow information (FAST3D)~\cite{fruhwirth2021fast3d}. Some works suggested multi-level self-supervised learning at global and local scales~\cite{fan2022self}. Recently, Cardace \textit{et al.}~\cite{cardace2023self} introduced a point cloud-specific self-training strategy with pseudo label refinement that exploits self-distillation to learn effective representations.

In addition to the self-training approaches, other approaches concentrate on establishing alignment between the source and target domains. Domain alignments are established through various strategies, including a multi-level alignment of local and global features (PointDAN)~\cite{qin2019pointdan}, a scale-aware and range-aware alignment strategy (SRDAN)~\cite{zhang2021srdan}, a student-teacher network along with pseudo-labeling (MLC-Net)~\cite{luo2021unsupervised}, and a contrastive co-training scheme (3D-CoCo)~\cite{yihan2021learning}. Shen \textit{et al.}~\cite{shen2022domain} introduced geometry-aware implicit in point clouds to mitigate domain biases and adopted a pseudo-labeling approach as part of their two-step method. CoSMix~\cite{saltori2022cosmix,saltori2023compositional} addresses domain shift in point cloud data through a compositional semantic mixup strategy with a teacher-student learning scheme.
Traditional UDA face limitations in scenarios where data privacy and data portability are critical. 
To address these constraints, source-free approaches facilitate domain adaptation exclusively using a source-trained model, eliminating the need for source domain labels~\cite{huang2021model}. 
Hegde and Patel~\cite{hegde2021attentive} utilize a transformer module to calculate an attentive class prototype, pinpointing accurate regions of interest for self-training. UAMT~\cite{hegde2021uncertainty} introduces a mean teacher approach with Monte-Carlo dropout uncertainty for supervising the student model using iteratively generated pseudo labels. In their subsequent work~\cite{hegde2023source}, the same authors propose an uncertainty-aware mean teacher framework that enhances conventional pseudo-label based on self-training methods by reducing the impact of label noise and assigning lower weights to samples when the teacher model is uncertain.
While UDA and SFDA approaches address critical challenges, they assume prior knowledge about test data distributions.

\noindent\textbf{Test-time domain adaptation:}
Test-time adaptation (TTA) is in contrast to traditional unsupervised domain adaptation, as it adapts a source-trained model for a new target domain during inference without using source data. 
A common method to minimize the domain difference when source data is unavailable is fine-tuning the source model using an unsupervised loss function derived from the target distribution.
TTT~\cite{sun2020test} updates model parameters in an online manner by applying a self-supervised proxy task
on the test data.
TENT~\cite{wang2021tent} updates trainable batch normalization parameters at test time by minimizing the entropy of model prediction. 
SHOT~\cite{liang2020we} also minimizes the expected prediction entropy derived from the output softmax distribution.
SHOT++~\cite{liang2021source} employs target-specific clustering for denoising pseudo labels.
A gradient-free TTT approach by Boudiaf \textit{et al.}~\cite{boudiaf2022parameter} emphasizes output prediction consistency while incorporating Laplacian regularization. 
TTT++~\cite{liu2021ttt++} introduces an extra self-supervised branch using contrastive learning in the source model to facilitate adaptation in the target domain.
TTT-MAE~\cite{gandelsman2022test}, an extension of TTT that utilizes a transformer backbone and replaces the self-supervision with masked autoencoders.
AdaContrast~\cite{chen2022contrastive} employs contrastive learning into TTA and utilizes both pseudo-label loss and diversity loss for optimization.
Finally, TTTFlow~\cite{osowiechi2023tttflow} employs unsupervised normalizing flows as an alternative to self-supervision for the auxiliary task.

\noindent\textbf{Test-time point cloud domain adaptation:}
While the TTA concept, originally designed for 2D image data, may encounter performance challenges when extended to 3D data, specialized designs are often required for 3D scenarios. However, a significant research gap remains in developing networks that can adapt a pre-trained source model during testing without requiring access to either source data or target labels.
MM-TTA~\cite{shin2022mm} introduces a selective fusion strategy to ensemble predictions from multiple modalities, enhancing self-learning signals in 3D semantic segmentation. It includes intra- and inter-modules for generating and selecting pseudo-labels. 

MATE~\cite{mirza2023mate} uses masked autoencoding as a robust self-supervised auxiliary objective to improve the network's resilience to distribution shifts in 3D point clouds.
DSS~\cite{Wang2023ContinualTD} is a method that addresses noisy pseudo-labels in TDA by proposing dynamic thresholding, positive learning, and negative learning processes. It demonstrates versatility for 3D point cloud classification by monitoring prediction confidence online and selecting low- and high-quality samples for training.
Hatem \textit{et al.}~\cite{hatem2023test} proposes a test-time adaptation approach for point cloud upsampling using meta-learning to enhance model generalization at inference time.
Point-TTA~\cite{hatem2023point} proposes a test-time adaptation approach for point cloud registration. The approach adapts the model parameters in an instance-specific manner during inference and obtains a different set of network parameters for each instance.
\section{Method}
\label{sec:method}

\begin{figure*}[!t]
\centering
\begin{center}
\includegraphics[width=0.98\textwidth]{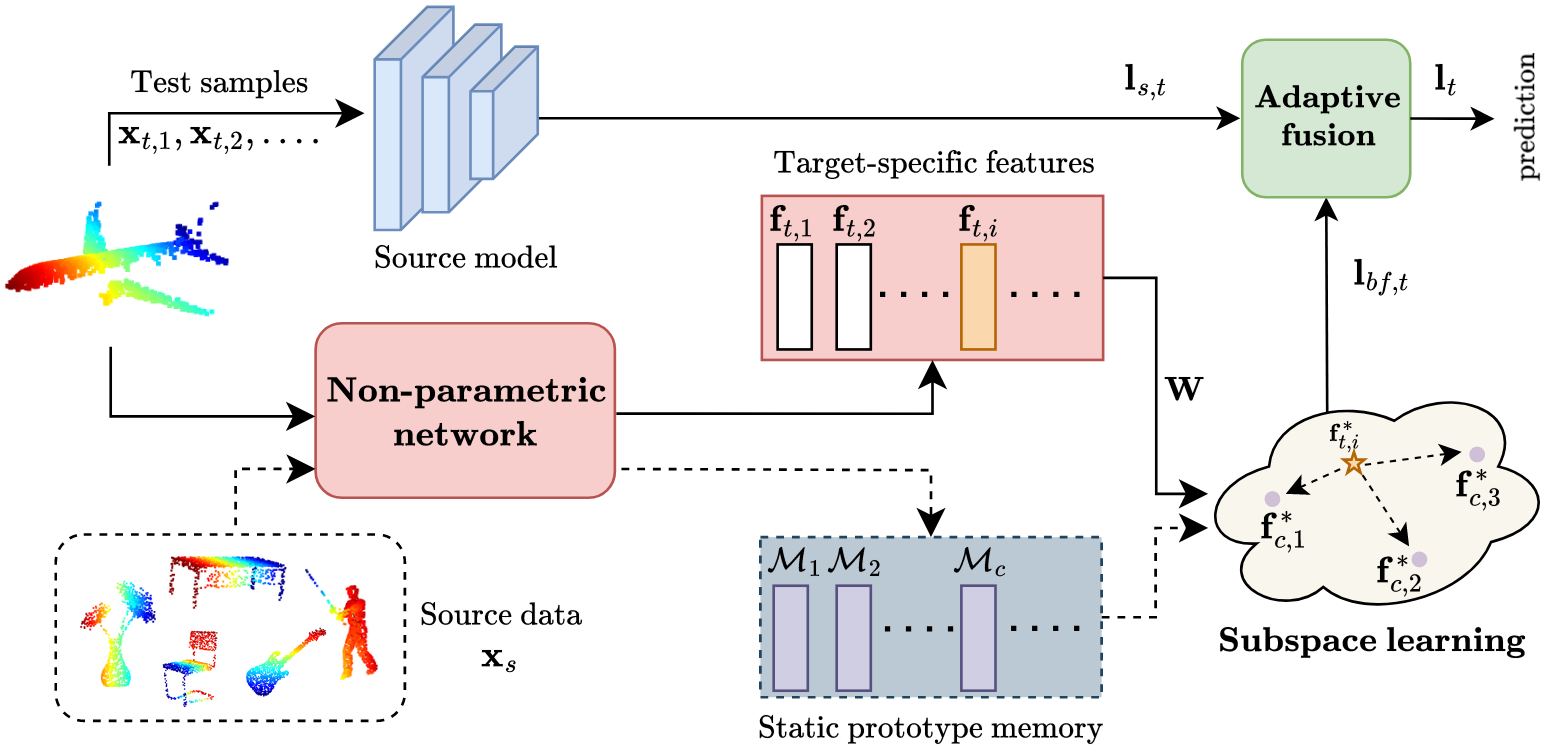}
\end{center}
\vspace{-8pt}
\caption{The framework of backpropagation-free test-time 3D model (BFTT3D). In the preparation stage, we first extract general features for source point cloud data $\mathbf{x}_s$ using a non-parametric network and then select a subset of all general features as static prototype memory $\mathcal{M}$. At test time, BFTT3D again adopts the non-parametric network to extract the general feature representation $\mathbf{f}_{t}$ from the given test point cloud sample $\mathbf{x}_t$ of $t$ domain. The feature $\mathbf{f}_{t}$ is then compared with static prototype feature $\mathbf{f}_c \in \mathcal{M}_c$ on a shared subspace to compute the target-specific logit $\mathbf{l}_{bf}$. Finally, the logit $\mathbf{l}_{bf}$ supplements the logit produced by the source model, $\mathbf{l}_{s}$, via an adaptive fusion module based on prediction entropy to output the final logit $\mathbf{l}_{t}$ for prediction. Notably, each module of BFTT3D, including the non-parametric network, subspace learning, and adaptive fusion, does not introduce any parameters that need backpropagation during adaptation.}
\label{model_overview}
\end{figure*}

\subsection{Problem Formulation}
In the context of Test-time Adaptation (TTA), we define two domains: the source domain denoted as $\mathcal{Q}_{s}$ and the target domain denoted as $\mathcal{Q}_{t}$. We assume the existence of a model, $h_{\theta_s}(.)$, representing the source model, where $\theta_s$ represents the pre-trained parameters on the source data of $\mathcal{Q}_{s}$. In the context of the 3D TTA task, the primary objective is for the source model $h_{\theta_{s}}(.)$ to accurately predict the point cloud sample $\mathbf{x}_t$ from the target domain $\mathcal{Q}_{t}$ during the test phase. To achieve this, the model undergoes adaptation to samples $\mathbf{x}_{t}$ from $\mathcal{Q}_{t}$ at test time. After this adaptation process, the updated model predicts the class label of $\mathbf{x}_{t}$. It is important to emphasize that, for reasons such as privacy or memory limitations, we do not have access to the source dataset $\mathcal{Q}_{s}$ during test time.
In this paper, following~\cite{choi2022improving}, we utilize a limited set of prototype feature representations from the source domain.

\subsection{Model Overview}
We provide a concise overview of our proposed method, BFTT3D, illustrated in Figure~\ref{model_overview}. Given test point cloud samples, $\{\mathbf{x}_{t,1}, \mathbf{x}_{t,2}, ..., \mathbf{x}_{t,n_{t}} \}$ from domain $\mathcal{Q}_{t}$, our goal is to predict class labels for these samples. 
The designed BFTT3D first extracts information using a non-parametric network that typically remains unaffected by domain bias, and thus it produces powerful target-specific features for the test 3D samples, $\{\mathbf{f}_{t,1}, \mathbf{f}_{t,2}, ..., \mathbf{f}_{t,n_{t}} \}$. These features will be compared with static prototypes in a shared subspace, and then the target-specific logit, $\mathbf{l}_{bf,t}$, will be computed. At last, $\mathbf{l}_{bf,t}$ complements the logit generated from the source model, $\mathbf{l}_{s,t}$, via an adaptive fusion module to output the final logit, $\mathbf{l}_{t}$, for prediction. It should be noted that modules we design along with the source model, including non-parametric network, subspace learning, and adaptive fusion, do not require any parameters that need backpropagation. 

\subsection{Backpropagation-free Adaptation} 
To better adapt the model to diverse domain distributions, we complement the source model logit $\mathbf{l}_{s,t}$ during test time of $\mathcal{Q}_{t}$. To achieve this, we generate the logit $\mathbf{l}_{bf,t}$ from the backpropagation-free adaptation module.
In particular, the non-parametric network transforms the incoming test 3D data $\mathbf{x}_{t}$ into the target-specific feature $\mathbf{f}_{t}$. 
Then, the target-specific logit $\mathbf{l}_{bf,t}$ is calculated based on $\mathbf{f}_{t}$ and the static prototype feature $\mathbf{f}_{c}$ in a common subspace. 

\subsubsection{Non-parametric Network}
Non-parametric point cloud network~\cite{zhang2023starting} does not require training and is constructed solely from non-learnable components, including farthest point sampling (FPS), k-nearest neighbors ($k$-NN), and pooling operations, supplemented by trigonometric functions. It has been demonstrated that the non-parametric network has a strong generalization ability from seen to unseen 3D data due to its training-free manner~\cite{zhang2023starting}. Given an input point cloud, $\mathbf{x}_{t,i}$ from $\mathcal{Q}_{t}$, trigonometric functions first encode the sample into the channel embedding with size $d$ for X, Y, and Z coordinates. For each channel s $ \in \{\text{X}, \text{Y}, \text{Z}\}$ , the channel embedding is computed as follows:
\begin{equation}
    g(\mathbf{x}_{t,i})^{(s)}=
    \begin{cases}
      \text{sin}(\alpha \mathbf{x}_{t,i}^{(s)}/\beta^{\frac{6i}{d}}), & i=2k \\
      \text{cos}(\alpha \mathbf{x}_{t,i}^{(s)}/\beta^{\frac{6i}{d}}),  & i=2k+1 \\
    \end{cases}
     \in \mathbb{R}^{d/3}
\end{equation}
where $\alpha$ and $\beta$ represent the wavelength and scale hyper-parameters. The raw point embedding $\mathbf{f}_{raw,t,i}$ is then calculated by the concatenated channel-wise embeddings from XYZ coordinates as follows:
\begin{equation}\label{raw_embedding}
\mathbf{f}_{raw,t,i}=[g(\mathbf{x}_{t,i})^{(\text{X})}, g(\mathbf{x}_{t,i})^{(\text{Y})}, g(\mathbf{x}_{t,i})^{(\text{Z})}] \in \mathbb{R}^{d}
\end{equation}

Given each point $\mathbf{x}$, we can obtain the embedding $\mathbf{f}_{raw}$. FPS is used to find the down-sampled local center points $\mathbf{x}_{cen}$ from all $\mathbf{x}$, and $k$-NN is then applied to group $\mathcal{N}_{ct}$ neighbor points $\mathbf{x}_{nb}$ for each center point $\mathbf{x}_{cen}$. To incorporate both center and neighbor information, we first calculate the expanded center feature $\bar{\mathbf{f}}_{cen}$ for $\mathbf{x}_{cen}$ by simply concatenating with neighboring point features:
\begin{equation}\label{Concat}
\bar{\mathbf{f}}_{cen} = \text{Concat}(\mathbf{f}_{cen}, \mathbf{f}_{nb,j}), j \in \mathcal{N}_{cen}
\end{equation}
where $\mathbf{f}_{cen}$ is the feature of point $\mathbf{x}_{cen}$ (See Eq.~(\ref{raw_embedding})), and 
$\mathbf{f}_{nb}$ denotes the feature of the neighbor point $\mathbf{x}_{nb}$. To show the spatial distribution of neighboring points $\mathbf{x}_{nb}$, the feature $\bar{\mathbf{f}}_{cen}$ is further reweighted by its relative position encoding $g(\Delta_{nb})$ to the center point:
\begin{equation}
    \bar{\mathbf{f}}_{cen}^w = (\bar{\mathbf{f}}_{cen} + g(\Delta_{nb})) \odot g(\Delta_{nb})
\end{equation}
where $\odot$ represents element-wise multiplication and $\Delta_{nb}$ is the relative position of $\mathbf{x}_{nb}$ to its center point $\mathbf{x}_{cen}$. Finally, the maximum pooling and average pooling are used to condense the information of $\bar{\mathbf{f}}_{cen}^w$:
\begin{equation}\label{MaxAvg}
    \bar{\mathbf{f}}_{cen}^p = \text{MaxPool}(\bar{\mathbf{f}}_{cen}^w)+\text{AvgPool}(\bar{\mathbf{f}}_{cen}^w)
\end{equation}

The process from Eq.~(\ref{Concat}) to (\ref{MaxAvg}) will be repeated four times. After the repetition is completed, a global pooling operation is applied to get the sample feature $\mathbf{f}_{t,i} \in \mathbb{R}^{d}$ for $\mathbf{x}_{t,i}$, which is shown in Figure~\ref{model_overview}.

\subsubsection{Subspace Learning}
\noindent\textbf{Prototype memory.}
Before test time, we use the non-parametric network to extract features from the source data $\mathbf{x}_s$ and save them in the static memory. To predict the logit, the non-parametric network uses similarity matching instead of the typical classification head. Hence, static memory should be employed to gather necessary category information from the source domain for achieving similarity matching~\cite{zhang2023starting}.

To be resource-efficient regarding memory usage, we employ the herding algorithm introduced by \cite{simon2021learning} combined with a nearest-mean-selection technique to select a subset from all features, ensuring that we only conserve limited memory space. We denote the whole feature set without selection for each class c as $\mathcal{M}_{c, all}$, and the selected exemplars set as $\mathcal{M}_{c}$. During the selection, for $\forall \mathbf{f} \in  \mathcal{M}_{c}$,
\begin{equation}
\operatorname{dist}\left(\mathbf{f}, \bar{\mathbf{f}}_c\right) \leq \min_{\mathbf{f}^{\prime} \in {\mathcal{M}_{c, all}}\backslash \mathcal{M}_{c}}  \operatorname{dist}\left(\bar{\mathbf{f}}_c, \mathbf{f}^{\prime}\right)
\end{equation}
where $\bar{\mathbf{f}}_c$ denotes mean averaged feature in $\mathcal{M}_{c, all}$. $\operatorname{dist(.,.)}$ is the distance function between two features. The selected features $\mathbf{f}_{c} \in \mathcal{M}_{c}$ are then stored in a fixed prototype memory.
The feature $\mathbf{f}_{t}$ from the non-parametric network and $\mathbf{f}_{c}$ from the prototype memory $\mathcal{M}_{c}$ share the same dimension.
To make a prediction, we construct the similarity matrix $\mathcal{J}$ between the feature vector $\mathbf{F}_t = [\mathbf{f}_{t,1}, \mathbf{f}_{t,2}, ..., \mathbf{f}_{t,n_{t}}]$ and the source prototype memory $\mathcal{M} = [\mathcal{M}_{1}, \mathcal{M}_{2}, ..., \mathcal{M}_{c}]$ via cosine similarity:
\begin{equation}\label{similarity_matrix}
   \mathcal{J} =  \tilde {\mathbf{F}}_{t} \cdot \tilde {\mathcal{M}}^{T}
\end{equation}
where $\tilde{\mathbf{F}}_{t}$ and $\tilde {\mathcal{M}}$ denote the corresponding normalized vectors of $\mathbf{F}_{t}$ and static memory $\mathcal{M}$.
Given the source memory label $\mathbf{L}_{m}$, we compute the output logit $\mathbf{l}_{bf,t}$ via:
\begin{equation}
   \mathbf{l}_{bf,t} = \varphi(\mathcal{J} \mathbf{L}_{m})
\end{equation}
where $\varphi(x) = \text{exp}(-\gamma(1-x))$ serves as an activation for prediction and $\gamma$ is a scaling hyperparameter. In this manner, the source prototypes contribute to the target-specific logit $\mathbf{l}_{bf,t}$.

\noindent \textbf{Similarity matrix.} Due to the distribution gap between the source and target domain, making direct predictions by similarity matrix $\mathcal{J}$ across two domains could be problematic. As such, before Eq.~(\ref{similarity_matrix}), we would like to have a projection function $\psi$ that first maps both static prototypes and test sample features, $\mathbf{f} \in \mathcal{M}$ and $\mathbf{f}_{t}$, into a shared subspace before calculating the similarity:
\begin{align}\label{transformed_vectors}
\mathbf{f}     &\leftarrow \mathbf{f}^{*}     = \psi(\mathbf{f})      \\ \nonumber
\mathbf{f}_{t} &\leftarrow \mathbf{f}_{t}^{*} = \psi(\mathbf{f}_{t}) 
\end{align}
Using the projection function $\psi(.)$, we aim to minimize the domain divergence between the prototype and test features within the shared subspace after mapping. Here, we use the Maximum Mean Discrepancy (MMD) distance~\cite{borgwardt2006integrating} to measure the statistical difference between the prototype and test domain. This kernel-based measure is distribution-free and can be applied without imposing strict requirements on the distribution.
\begin{equation}\label{dist_source_test}
\resizebox{0.41\textwidth}{!}{
$
\operatorname{Dist}\left(\mathcal{Q}_s, \mathcal{Q}_t\right)=\left\|\frac{1}{n_s} \sum_{i=1}^{n_s} \mathbf{f}_{i}^{*}-\frac{1}{n_t} \sum_{i=1}^{n_t} \mathbf{f}_{t,i}^{*}\right\|_{\mathcal{H}}^2
$
}
\end{equation}
where $n_s$ is the number of features $\mathbf{f}$ in $\mathcal{M}$, and $\left\| \cdot \right\|_{\mathcal{H}}^2$  is the $l_2$ norm computed in a reproducing kernel Hilbert space. 

To approximate the mapping function $\psi(.)$ in practice, which can mitigate the difference between these static source prototypes and test sample features in the common space, we employ a classical transfer learning technique named Transfer Component Analysis (TCA)~\cite{pan2010domain}. In general, TCA finds a transformation matrix $\mathbf{W} \in \mathbb{R}^{\left(n_s+n_t\right) \times m}$, as the empirical measurement of mapping function $\psi$ induced by a universal kernel, to project both $\mathbf{f}_s$ and $\mathbf{f}_t$ onto a commonly shared subspace of dimension $m$. The objective is to minimize the MMD distance on the mapped subspace formulated in a kernel learning problem~\cite{borgwardt2006integrating}. The source-target kernel matrix $\mathbf{K}$ is formulated as follows, reflecting data similarities in the source, target, and across domains.
\begin{equation}
\mathbf{K} =\left\langle\psi\left(\mathbf{x}_i\right), \psi\left(\mathbf{x}_j\right)\right\rangle_{i, j=1}^{n_s+n_t} \in \mathbb{R}^{(n_s+n_t) \times (n_s+n_t)}    
\end{equation}
\begin{equation}
\mathbf{L}_{i, j}= \begin{cases}\frac{1}{n_t^2} & \text { if } \mathbf{x}_i, \mathbf{x}_j \in \mathcal{Q}_t \\ \frac{1}{n_t^s} & \text { if } \mathbf{x}_i, \mathbf{x}_j\in \mathcal{Q}_s \\ \frac{-1}{n_{s}n_{t}} & \text{otherwise}\end{cases},
\end{equation}
where $\left\langle \cdot \right\rangle$ denotes the dot product in $\mathcal{H}$, and $\mathbf{L}$ is used to scale the kernel matrix $\mathbf{K}$. The overall kernel learning objective for MMD distance can be rewritten as follows, according to Eq.~(\ref{dist_source_test}):
\begin{equation}
\begin{aligned}
\min _{\mathbf{W} \in \mathbb{R}^{(n_s+n_t) \times m}} & \operatorname{tr}\left(\mathbf{W}^{T} \mathbf{K} \mathbf{L} \mathbf{K} \mathbf{W}\right)+\mu \operatorname{tr}\left(\mathbf{W}^{T} \mathbf{W}\right), \\
\text { s.t. } & \mathbf{W}^{T} \mathbf{K} \mathbf{H} \mathbf{K} \mathbf{W}=\mathbf{I}
\end{aligned}
\end{equation}
where $\mu$ represents the regularization penalty to control the complexity of $\mathbf{W}$ and $\mathbf{H}=\mathbf{I}-\mathbf{1 1}^{T} /\left(n_s+n_t\right) \in \mathbb{R}^{\left(n_s+n_t\right) \times\left(n_s+n_t\right)}$ is the centering matrix. By solving this constrained trace optimization problem with regularization, the solution of $\mathbf{W}$ is explicitly given by the $m$ largest eigenvectors of 
\begin{equation}
\begin{aligned}(\mathbf{K L K}+\mu \mathbf{I})^{-1} \mathbf{K} \mathbf{H K}.\end{aligned}
\end{equation} 

Once $\mathbf{W}$ is obtained, the transformed source prototype and target-specific feature vectors, $\mathbf{f}^{*}$ and $\mathbf{f}_{t}^{*}$ in Eq.~(\ref{transformed_vectors}), are the corresponding indexed columns of $\mathbf{W}^{T} \mathbf{K} \in \mathbb{R}^{m \times \left(n_s+n_t\right)}$.

\subsection{Adaptive Fusion Module}
After obtaining the target-specific logit $\mathbf{l}_{bf,t}$ from the backpropagation-free adaptation, as shown in Figure~\ref{model_overview}, we then fuse $\mathbf{l}_{bf,t}$ and the source-specific logit $\mathbf{l}_{s,t}$ from the source model, into a final logit $\mathbf{l}_{t}$. 
The distribution distance between the source and target domain varies since the target domain is diverse.
As such, to tackle the varying distribution gaps, we adaptively fuse $\mathbf{l}_{s,t}$ and $\mathbf{l}_{bf,t}$ for a more accurate prediction. When the incoming test point cloud samples are drawn from a target domain whose distribution is similar to the source data, we hope to emphasize a larger proportion of the logit from the source model branch since it is reliable in predicting in-distribution test samples. Conversely, when the domain sample is drawn from a significantly different distribution than the source domain, we prioritize using a larger proportion of the logit from the backpropagation-free adaptation to provide more target-specific information. To achieve this, we first adopt a weight $p$ to control the proportion of the logit used in the final prediction.
\begin{equation}
\mathbf{l}_{t} = p \cdot \mathbf{l}_{bf,t} + (1-p)\cdot \mathbf{l}_{s,t}
\end{equation}
To dynamically calculate the value of $p$ in the test time, we employ the softmax entropy of logits as the measurement to give $p$ since entropy is related to the errors and shifts~\cite{wang2021tent}. Intuitively, entropy is related to the model's certainty about the test point cloud samples, as low-entropy predictions are most likely correct with high confidence. Thus, we calculate the entropy ratio between two logits as follows to reflect the degree of the difference between the current target and source domain:
\begin{equation}\label{p}
p = 1 - \dfrac{\text{E}(\mathbf{l}_{bf,t})}{\text{E}(\mathbf{l}_{s,t}) + \text{E}(\mathbf{l}_{bf,t})}
\end{equation}
where $\text{E}(.)$ denote the entropy. In this way, we achieve the adaptive fusion of the logits from the source model and backpropagation-free adaptation module.
\section{Experiments}
\label{sec:exp}
In this section, we provide extensive experiments to demonstrate the effectiveness of the proposed BFTT3D model in TTA for 3D data. We evaluate our method on multiple 3D point cloud datasets, including ModelNet-40C~\cite{sun2022benchmarking} and ScanObjectNN-C~\cite{mirza2023mate}. 

\begin{table*}[t]
\centering
\resizebox{1.0\textwidth}{!}{
\begin{tabular}{l|c|c|c|c|c|c|c|c|c|c|c|c|c|c|c|c|c}
\toprule
\textbf{Method} &\textbf{Backbone} &\rotatebox{0}{\textbf{uniform}} &\rotatebox{0}{\textbf{gaussian}} &\rotatebox{0}{\textbf{background}} & \rotatebox{0}{\textbf{impulse}} &\rotatebox{0}{\textbf{upsampling}} & \rotatebox{0}{\textbf{rbf}} &\rotatebox{0}{\textbf{rbf-inv}} &\rotatebox{0}{\textbf{den-dec}} &\rotatebox{0}{\textbf{dens-inc}} & \rotatebox{0}{\textbf{shear}} & \rotatebox{0}{\textbf{rot}} & \rotatebox{0}{\textbf{cut}} &\rotatebox{0}{\textbf{distort}}&\rotatebox{0}{\textbf{oclsion}} & \rotatebox{0}{\textbf{lidar}} & \textbf{Mean} $\downarrow$\\
\bottomrule

\textbf{Source-only}  &\multirow{7}*{\textbf{PointNet}~\cite{qi2017pointnet}}
                                    &14.63&18.68&95.30&33.39&15.03&29.46&27.63&12.93&10.49&42.71&72.81&14.95&34.85&56.28&59.00&35.88 \\
\textbf{TENT~\cite{wang2021tent}}   &   &14.71&18.35&90.36&27.03&15.36&28.28&26.66&13.86&12.36&40.68&65.68&15.76&33.87&56.56&59.85&34.62 \\
\textbf{BN~\cite{li2016revisiting}} &   &14.83&18.84&89.63&27.47&15.15&28.69&26.66&14.59&12.40&40.44&65.88&16.25&34.68&57.58&60.21&34.89 \\
\textbf{SHOT~\cite{liang2020we}}    &   &14.91&17.22&90.44&25.61&14.51&27.07&25.85&13.78&12.20&39.95&65.80&16.21&32.98&56.93&60.29&34.25 \\

\textbf{LAME~\cite{boudiaf2022parameter}}  &&76.64&84.67&95.95&95.48&67.83&93.37&91.72&59.92&34.81&96.45&97.50&68.52&92.53&95.99&95.95&83.16 \\
\textbf{DSS~\cite{Wang2023ContinualTD}}&&14.20&18.71&94.13&33.12&14.33&29.31&27.55&13.31&10.92&41.42&66.31&15.45&34.10&56.31&59.33&35.23 \\

\rowcolor{gray}
\textbf{BFTT3D (ours)}  &   &14.55&18.27&80.75&31.93&14.83&28.97&27.19&12.76&10.49&39.71&68.56&14.67&34.04&54.13&55.88&\textbf{33.78} \\
\hline

\textbf{Source-only}  &\multirow{5}*{\textbf{DGCNN}~\cite{dgcnn}}  
                                         &23.46&28.20&57.41&37.93&33.23&22.73&20.91&27.59&16.57&15.96&41.33&23.78&19.94&65.52&85.25&34.65 \\
\textbf{TENT~\cite{wang2021tent}}    &   &18.92&19.94&61.91&22.85&24.64&21.80&20.87&25.00&18.15&18.76&31.97&23.46&21.72&62.56&72.29&30.99 \\
\textbf{BN~\cite{li2016revisiting}}  &   &19.81&20.42&63.01&23.22&25.41&22.41&21.52&26.09&17.54&18.80&32.09&23.10&21.76&63.13&72.12&31.36 \\
\textbf{SHOT~\cite{liang2020we}}     &   &18.88&19.73&57.13&21.27&22.77&22.29&19.73&24.31&17.59&18.48&31.28&22.12&21.72&62.28&72.97&30.17 \\

\rowcolor{gray}
\textbf{BFTT3D (ours)}               &   &19.45&20.02&58.51&22.33&24.59&21.96&20.71&24.03&16.49&18.80&31.48&21.64&21.39&56.28&62.24&\textbf{29.33} \\
\hline

\textbf{Source-only}   &\multirow{5}*{\textbf{Curvenet}~\cite{xiang2021walk}}   
                                          &15.60&17.42&81.24&35.49&12.80&14.22&12.88&20.87&10.90&12.32&29.58&20.38&12.93&62.12&68.35&28.47 \\
\textbf{TENT~\cite{wang2021tent}}     &   &13.94&13.01&56.20&20.06&12.76&13.45&12.72&15.15&10.90&12.32&19.57&16.65&13.70&53.57&55.06&22.60 \\
\textbf{BN~\cite{li2016revisiting}}   &   &14.95&14.59&60.66&21.35&13.05&15.07&13.61&17.06&11.43&13.65&21.27&17.75&14.10&55.15&56.77&24.03 \\
\textbf{SHOT~\cite{liang2020we}}      &   &11.75&11.35&43.01&16.17&10.58&11.30&10.94&11.75&9.44&10.01&16.03&12.28&12.75&55.19&54.28&19.79\\

\rowcolor{gray}
\textbf{BFTT3D (ours)} &              &11.87&12.16&40.68&15.36&10.74&11.51&11.02&11.99&10.01&10.72&16.82&13.74&11.75&50.32&49.88&\textbf{19.23} \\
\hline
\end{tabular}}
\caption{Experimental results on ModelNet-40C~\cite{sun2022benchmarking}. The classification errors in \% are provided.}
\label{tab:Modelnet40-c-backbone}
\end{table*}

\begin{table*}[t]
\centering
\resizebox{1.0\textwidth}{!}{
\begin{tabular}{l|c|c|c|c|c|c|c|c|c|c|c|c|c|c|c|c|c}
\toprule
\textbf{Method} &\textbf{Backbone} &\rotatebox{0}{\textbf{uniform}} &\rotatebox{0}{\textbf{gaussian}} &\rotatebox{0}{\textbf{background}} & \rotatebox{0}{\textbf{impulse}} &\rotatebox{0}{\textbf{upsampling}} & \rotatebox{0}{\textbf{rbf}} &\rotatebox{0}{\textbf{rbf-inv}} &\rotatebox{0}{\textbf{den-dec}} &\rotatebox{0}{\textbf{dens-inc}} & \rotatebox{0}{\textbf{shear}} & \rotatebox{0}{\textbf{rot}} & \rotatebox{0}{\textbf{cut}} &\rotatebox{0}{\textbf{distort}}&\rotatebox{0}{\textbf{oclsion}} & \rotatebox{0}{\textbf{lidar}} & \textbf{Mean} $\downarrow$\\
\bottomrule

\textbf{Source-only}  &\multirow{7}*{\textbf{PointNet}~\cite{qi2017pointnet}}
                                    &63.17&40.62&87.61&70.57&62.31&60.24&58.52&23.24&22.89&65.23&69.54&26.51&60.59&89.33&89.85&59.35 \\
\textbf{TENT~\cite{wang2021tent}}   &   &61.10&38.55&84.34&69.36&60.24&58.69&57.14&23.24&22.72&63.51&68.16&25.82&59.55&89.50&90.71&58.18 \\
\textbf{BN~\cite{li2016revisiting}} &   &60.76&37.01&83.65&68.16&59.21&58.86&57.14&24.27&23.58&63.34&67.99&25.82&58.18&89.5&90.53&57.87 \\
\textbf{SHOT~\cite{liang2020we}}    &   &62.20&55.77&88.64&86.57&63.86&70.91&67.13&27.54&25.47&74.01&75.39&28.23&68.33&93.29&95.18&65.63 \\

\textbf{LAME~\cite{boudiaf2022parameter}} &&60.24&33.31&91.91&70.22&59.71&59.90&54.77&14.53&15.98&55.52&62.90&16.44&58.04&92.73&88.92&55.67  \\
\textbf{DSS~\cite{Wang2023ContinualTD}}&&59.83&34.74&84.81&69.26&58.21&57.13&55.42&21.69&20.11&61.52&65.53&22.45&58.21&88.47&90.53&56.53 \\

\rowcolor{gray}
\textbf{BFTT3D (ours)}  &
&57.14&39.93&70.05&65.23&55.94&54.56&53.01&21.34&21.51&58.69&68.85&24.10&55.94&85.03&85.54&\textbf{54.46}\\
\hline

\textbf{Source-only}  &\multirow{5}*{\textbf{DGCNN}~\cite{dgcnn}}  
                                         &57.83&49.91&55.94&43.37&53.53&37.01&34.08&24.10&19.62&33.39&38.73&25.13&34.60&92.08&92.08&46.09 \\
\textbf{TENT~\cite{wang2021tent}}    &   & 54.73&47.16&55.25&43.89&52.84&36.83&32.53&24.78&20.48&31.50&38.55&23.92&32.53&91.39&91.22&45.17 \\
\textbf{BN~\cite{li2016revisiting}}  &   &55.25&47.16&55.42&42.86&52.84&36.32&32.19&24.78&19.45&32.01&38.38&24.61&32.36&91.39&91.57&45.11 \\
\textbf{SHOT~\cite{liang2020we}}     &   &86.92&67.81&81.93&65.75&86.75&43.20&37.01&30.46&27.19&36.32&46.13&32.53&40.45&91.22&89.33&57.53 \\
\rowcolor{gray}
\textbf{BFTT3D (ours)}               &   &57.31&41.82&52.32&43.37&53.36&36.83&34.07&22.55&17.73&33.05&38.55&22.20&34.58&85.54&85.37&\textbf{43.91} \\
\hline

\textbf{Source-only}  &\multirow{5}*{\textbf{Curvenet}~\cite{xiang2021walk}}  
                                         &69.54&52.32&81.93&80.21&67.99&67.13&66.27&35.46&31.15&64.03&69.36&36.32&67.13&90.53&92.25&64.77 \\
\textbf{TENT~\cite{wang2021tent}}    &   &70.22&53.87&81.93&80.21&70.05&68.16&67.81&36.49&32.53&66.09&70.57&36.83&67.99&91.05&92.77&65.77 \\
\textbf{BN~\cite{li2016revisiting}}  &   &69.36&50.95&80.21&79.52&68.16&66.95&66.09&34.77&31.33&64.37&67.99&35.63&66.09&90.36&92.08&64.26 \\
\textbf{SHOT~\cite{liang2020we}}     &   &87.26&77.11&91.91&93.98&87.09&87.78&87.44&72.12&73.84&87.26&87.44&74.18&87.09&95.01&90.71&85.35 \\
\rowcolor{gray}
\textbf{BFTT3D (ours)}               &   &61.96&48.19&68.85&71.43&60.93&58.00&57.83&30.29&28.23&57.49&62.65&31.50&58.86&88.64&85.54&\textbf{57.80} \\

\hline

\end{tabular}}
\caption{Experimental results on ScanObjectNN-C~\cite{mirza2023mate}. The classification errors in \% are provided.}
\label{tab:Scanobjectnn-c-backbone}
\end{table*}

\subsection{Implementation} 
For the non-parametric network, we follow the implementations in \cite{zhang2023starting}. Regarding the static prototype memory, we construct the feature memory only once before adaptation and keep it fixed throughout the adaptation stage. We selectively choose $25\%$ of the samples as the prototype memory by herding algorithm, and the transformed feature dimension $m$ for subspace learning is fixed to $150$. Note that the severity of corruption applied is fixed to the highest $5$ for all test experiments. In addition, we also utilized various backbones for the source model, each trained independently on its corresponding clean set, to validate the robustness of BFTT3D. The number of points for the input point cloud is set to 1024. We run all experiments on a single NVIDIA RTX 4080.

\subsection{Datasets}
\noindent\textbf{ModelNet-40C.} ModelNet-40C~\cite{sun2022benchmarking} is a benchmark with 15 common types of corruptions
induced on the original test set of ModelNet-40~\cite{wu20153d}. The goal of building the ModelNet-40C dataset is to mimic distribution shifts that occur in the real world. 

\noindent\textbf{ScanObjectNN-C.} ScanObjectNN~\cite{uy-scanobjectnn-iccv19} is a point cloud classification dataset collected from the real world. It contains 15 classes, with 2309 samples in the train set and 581 in the test set. To build ScanObjectNN-C, we employ the setting proposed by~\cite{mirza2023mate} to generate 15 corruptions in the test set of ScanObjectNN with the object only.

\subsection{Main Results}
\noindent\textbf{ModelNet-40C.}
We first examine the adaptation performance on ModelNet-40C, and the experimental results are given in Table~\ref{tab:Modelnet40-c-backbone}. As shown in the table, all baseline methods, including TENT~\cite{wang2021tent}, BN~\cite{li2016revisiting}, and SHOT~\cite{liang2020we} have improvements over the source model. However, our method, BFTT3D, has the lowest error rate on average. Interestingly, our approach consistently demonstrates significant improvements in domains like \textit{occlusion} and \textit{lidar}. For example, for the target domain of \textit{occlusion}, BFTT3D achieves improvements of $5.02\%$, $9.24\%$, and $11.82\%$ over the source model. This observation suggests that our method is particularly effective when the target and source domains have a large gap.

\noindent\textbf{ScanObjectNN-C.}
The experimental result of 3D TTA performance on ScanObjectNN-C~\cite{mirza2023mate}, a more difficult case, are reported in Table~\ref{tab:Scanobjectnn-c-backbone}. Notably, the source model exhibits a high error rate for each backbone, indicating substantial domain divergence on average and increased difficulty in adaptation. Baseline methods show limited improvement in adapting the source model to test the domain. However, our BFTT3D model can still overcome the large domain gap. 
The table shows that our method has $4.89\%$, $2.18\%$, and $6.59\%$ less men error compared to the source model, using PointNet, DGCNN, and Curvenet as the backbone, respectively. 
This demonstrates that our approach provides useful target-specific information to the source model, even when the test target domain distribution differs largely from the source domain. Another fact is that most baseline methods face a limitation of insufficient test data for adaptation, particularly because ScanObjectNN-C has a very limited number of test samples for each target domain. However, this constraint does not affect BFTT3D as our method does not require samples for training the source model. It only involves concatenating the logits from the frozen source model and the backpropagation-free adaptation, offering greater flexibility.

\subsection{Ablation Study}
\noindent\textbf{Number of prototypes.}
Here, we explore the influence of the number of exemplars utilized for constructing the similarity matrix. The results in Table~\ref{tab:ablation-prototype} indicate that solely storing $100\%$ of source features for similarity construction yields limited benefits compared to using only $25\%$ of the data. Our setting of $25\%$ not only conserves less memory space but also reduces the size of the similarity matrix to improve processing speed.

\begin{table}[t]
\centering
\resizebox{0.28\textwidth}{!}{
\begin{tabular}{l|c|c}
\toprule
\textbf{Method} &\textbf{Ratio} &\textbf{Mean} $\downarrow$\\
\midrule
\textbf{Source}          &None               &46.09 \\
\textbf{BFTT3D}          &100\%              &\textbf{43.75} \\
\textbf{BFTT3D}          &75\%               &43.80 \\
\textbf{BFTT3D}          &50\%               &43.91 \\
\textbf{BFTT3D}          &25\%               &43.91 \\
\bottomrule
\end{tabular}}
\caption{Ablation study: number of prototype. The experiments run on ScanObjectNN-C~\cite{mirza2023mate}. The mean classification errors in \% are provided. The ratio reflects the proportion of the prototype memory relative to the total static memory, indicating the size of the prototype memory.}
\label{tab:ablation-prototype}
\end{table}

\noindent\textbf{Subspace learning method.}
As shown in Table~\ref{tab:ablation-subspace}, we present the performance of alternative subspace learning methods for creating shared subspace of the similarity matrix. The results show that certain subspace learning approaches for domain alignment, such as TCA~\cite{pan2010domain} and JDA~\cite{long2013transfer},  have relatively low errors compared to other setups. Others like CORAL~\cite{sun2016return} do not demonstrate effective improvements in performance. Overall, we note that the highest error occurs when no subspace learning is applied. This indicates that transforming onto a shared subspace before computing the similarity matrix in non-parametric adaptation is necessary.

\begin{table}[t]
\centering
\resizebox{0.28\textwidth}{!}{
\begin{tabular}{l|c}
\toprule
\textbf{Subspace method} & \textbf{Mean} $\downarrow$\\
\midrule
\textbf{None}                            & 56.84\\
\textbf{JDA}~\cite{long2013transfer}     & 54.87\\
\textbf{CORAL}~\cite{sun2016return}      & 56.80\\
\textbf{TCA}~\cite{pan2010domain}        &\textbf{54.46} \\
\bottomrule
\end{tabular}}
\caption{Ablation study: subspace learning method. The experiments run on ScanObjectNN-C~\cite{mirza2023mate}. Using our BFTT3D with PointNet backbone, different subspace learning methods are tested, including JDA, CORAL, and TCA. The mean classification errors in \% are provided.}
\label{tab:ablation-subspace}
\vspace{-2mm}
\end{table}

\begin{figure}[ht]
    \centering
    \includegraphics[width=0.45\textwidth]{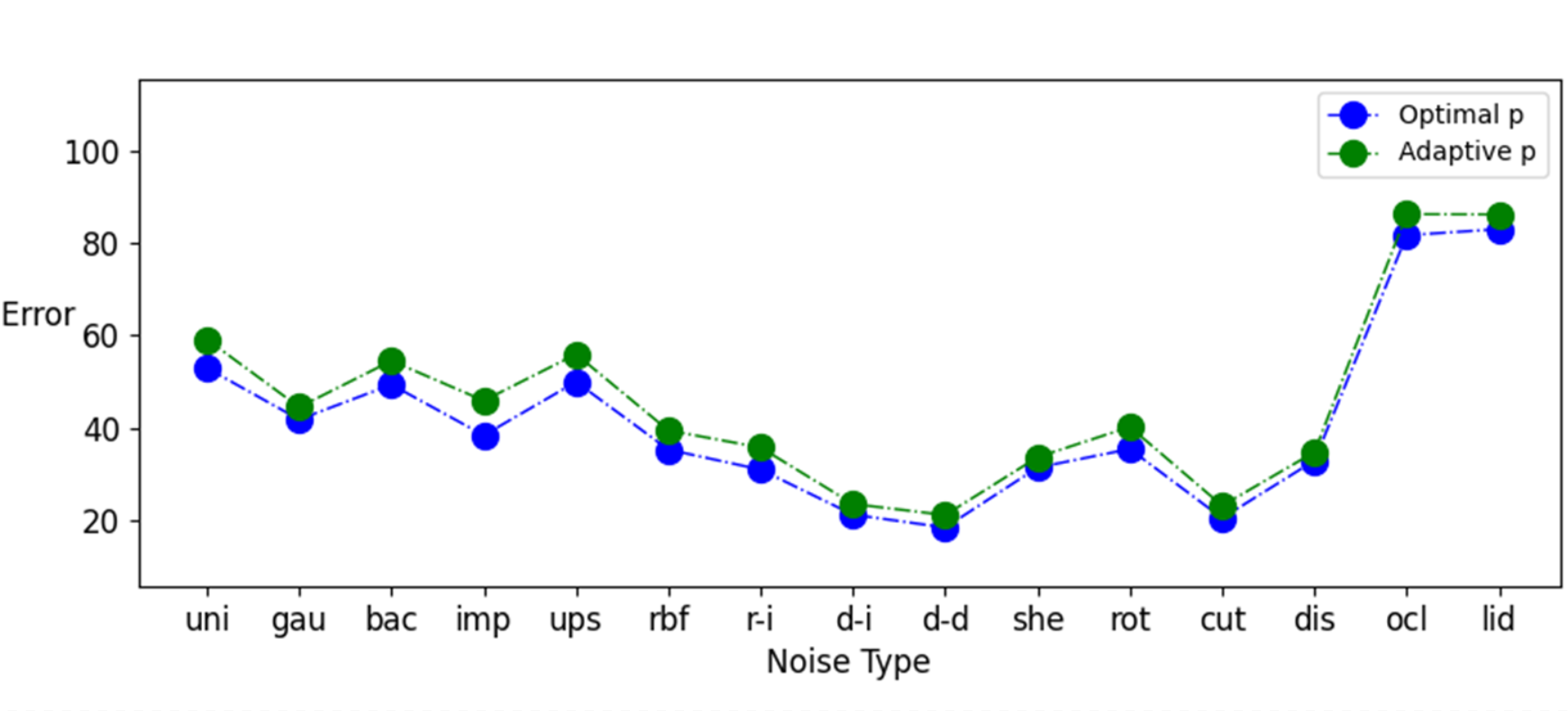}
    \caption{Comparisons of mean error between adaptive and optimal ratio. Using the backbone PointNet~\cite{qi2017pointnet} and testing on ModelNet40-C~\cite{sun2022benchmarking}. The blue line represents the optimal $p$ from an exhaustive search, and the green line represents our adaptive ratio $p$. 
    }
    \label{fig:adaptive_ratio_2}
\end{figure}

\noindent\textbf{Adaptive ratio.}
In this section, we first evaluate the effectiveness of the entropy-based ratio from the adaptive fusion module in comparison with fixed thresholds. As seen from Table~\ref{tab:ablation-ratio}, simply using $p=0.5$ to mix logits from the target-specific branch and source model helps to reduce the error compared with the source-only setting. By leveraging the adaptive ratio, we get the lowest mean error across the board compared with other setups. This may indicate that the adaptive ratio reaches a good balance between the two branches of logits. In addition,  we further conducted experiments on ModelNet40-C~\cite{sun2022benchmarking} to compare the adaptive ratio with a best $p$ found by exhaustive search to show the reliability of our adaptive ratio. Note that from Figure~\ref{fig:adaptive_ratio_2}, the error rate of the adaptive threshold is very similar to the error with the optimal $p$, staying within a certain small range. Moreover, it is unlikely for us to manually set an optimal and reliable threshold without validation. In contrast, our adaptive ratio provides a good estimated mixing threshold for test data, and we can calculate this without any validation in test time. 

\begin{table}[t]
\centering
\resizebox{0.32\textwidth}{!}{
\begin{tabular}{l|c|c}
\toprule
\textbf{Method} &\textbf{Ratio} &\textbf{Mean} $\downarrow$ \\
\midrule
\textbf{Source-only}  & None               &46.09 \\
\textbf{BFTT3D}       & $p = 0.5$          &44.13 \\
\textbf{BFTT3D}       & adaptive $p$       &\textbf{43.91} \\
\bottomrule
\end{tabular}}
\caption{Ablation study: adaptive ratio. Using the backbone CurveNet, the experiments run on ScanObjectNN-C~\cite{mirza2023mate}. The mean classification errors in \% are provided.}
\label{tab:ablation-ratio}
\end{table}

\vspace{-5mm}
\section{Conclusion}
In this work, we propose the model BFTT3D for the task of 3D TTA. In particular, BFTT3D first generates the complementary target-domain-specific logit via similarity matching on shared subspace. Then, it is combined with the source-domain-specific model logit via an entropy-based adaptive fusion strategy to output final refined predictions for test samples from diverse distributions. The entire framework of BFTT3D does not introduce any parameters that need backpropagation, thus avoiding the complex pseudo-labeling process. For future work, meta-training in the pre-training stage, for the simplicity of our method, could be a promising direction.

\noindent\textbf{Acknowledgment.} Mehrtash Harandi expresses gratitude for the support provided by the Australian Research Council via the Discovery Program (DP230101176).

{
 \bibliographystyle{ieeenat_fullname}
 \bibliography{main}
}

\end{document}